%% file: main.tex
\documentclass[letterpaper, 10 pt, conference]{ieeeconf}  

\usepackage{graphicx}
\usepackage{caption}
\usepackage{booktabs}
\usepackage{multirow}
\usepackage{xcolor}
\usepackage{siunitx}
\usepackage{pifont}
\usepackage{makecell}
\usepackage{tabularx}
\let\labelindent\relax
\usepackage{enumitem}
\usepackage[colorlinks=true, linkcolor=blue]{hyperref}
\usepackage{amssymb}
\usepackage{amsmath}
\usepackage{subcaption}

\IEEEoverridecommandlockouts                              

\title{\LARGE \bf
GaussianMap: Learning Gaussian Representation for \\ Multi-Sensor Online HD Map Construction
}

\author{
Hongyu Lyu\textsuperscript{1}, Julie Stephany Berrio Perez\textsuperscript{2}, Mao Shan\textsuperscript{1}, Stewart Worrall\textsuperscript{1} 
\thanks{\textsuperscript{1} The University of Sydney, Australian Centre for Robotics, Sydney, Australia. Emails: \texttt{\small \{h.lyu, m.shan, s.worrall\}@acfr.usyd.edu.au}}
\thanks{\textsuperscript{2} The University of Queensland, Australia. Email: \texttt{\small s.berrioperez@uq.edu.au}}
}

\begin{document}

\maketitle
\thispagestyle{empty}
\pagestyle{empty}

\begin{abstract}

Autonomous driving systems benefit from high-definition (HD) maps that provide critical information about road infrastructure. The online construction of HD maps offers a scalable approach to generate local vectorized maps from onboard sensor observations. Existing methods commonly adopt bird’s-eye-view (BEV) features as the intermediate scene representation, encoding the surrounding space with fixed-resolution dense grids. However, map elements are spatially sparse yet require fine-grained geometric localization, making uniformly allocated BEV representations redundant and less effective for vectorized map prediction. In this work, we propose GaussianMap, an online HD map construction framework that learns an adaptive Gaussian representation of the surrounding scene. This representation consists of a set of Gaussian primitives on the BEV plane, each encoding a flexible local region with geometric properties and a feature vector, allowing the model to allocate representational capacity to map-relevant regions. To generate such a representation from sensor observations, we introduce a feed-forward Gaussian encoder that progressively refines these primitives through Gaussian interaction modeling and multi-sensor feature aggregation. The refined Gaussian representation is then splatted into a BEV feature map and decoded into vectorized map predictions. Extensive experiments on nuScenes and Argoverse~2 datasets demonstrate that GaussianMap achieves state-of-the-art performance in both camera-only and camera-LiDAR fusion settings. Our code will be made publicly available.

\end{abstract}

\input{1_introduction}
\input{2_related_work}

\input{3_method}
\input{4_experiments}
\input{5_conclusion}






\bibliographystyle{IEEEtran}
\bibliography{main}

\end{document}

%% file: 1_introduction.tex
\section{Introduction}

High-Definition (HD) maps are an essential component of autonomous driving systems, as they provide accurate representations of the road infrastructure for safe and reliable navigation~\cite{lyu2025online}. These maps encode a wide range of road elements in a vectorized form, such as road boundaries, lane dividers, and pedestrian crossings~\cite{elghazaly2023high}. Conventionally, HD maps are constructed offline from multimodal data collected by survey vehicles, typically through SLAM-based pipelines~\cite{zhang2014loam, shan2018lego} followed by manual annotation and refinement. Although effective, this process is labor-intensive and costly, making map updates difficult in dynamic road environments~\cite{berrio2021long}. Therefore, the online construction of HD maps has attracted increasing attention, with the aim of generating map representations around the ego vehicle at runtime from onboard sensor observations~\cite{li2022hdmapnet, liu2023vectormapnet, liao2022maptr, lyu2025maprf}.

Despite recent progress in online HD map construction, building an effective and efficient representation of the surrounding scene remains a key challenge. Existing methods~\cite{liao2024maptrv2, liu2024leveraging, dong2025damap} commonly adopt bird's-eye-view (BEV) features as the intermediate scene representation. Although BEV features provide a structured top-down space for map prediction, they are typically built upon fixed-resolution dense grids, which are not well suited to the sparsity and geometric sensitivity of HD maps. Map elements usually occupy only a small fraction of the scene, yet their accurate prediction relies on fine-grained geometric details~\cite{lyu2025online, elghazaly2023high}. However, BEV grids allocate representational capacity uniformly across the perception range, including regions with limited map-relevant information. This uniform allocation leads to redundancy and limits the model's ability to focus on geometrically informative regions.

\input{figures/3_teaser}

To address these challenges, we propose GaussianMap, an online HD map construction framework that adaptively allocates representational capacity to map-relevant regions. As illustrated in Fig.~\ref{fig:teaser}, GaussianMap represents the surrounding scene with a set of Gaussian primitives on the BEV plane. Each Gaussian corresponds to a flexible local region and is parameterized by a mean, covariance, opacity, and feature vector. Unlike 3D Gaussian Splatting~\cite{kerbl20233d}, which optimizes the Gaussian parameters offline for each scene, we design a feed-forward Gaussian encoder that learns to generate a Gaussian representation of the scene from onboard sensor observations. This representation is then splatted into BEV features and fed into a vectorized map decoder.

To effectively learn such a representation, the Gaussian encoder progressively updates Gaussian primitives through feature interaction and property refinement. We introduce Gaussian self-attention to model interactions among Gaussians based on their spatial distributions, while Gaussian-camera attention incorporates visual information from multi-camera image features. When LiDAR is available, we design Gaussian-LiDAR attention to leverage complementary geometric cues from LiDAR BEV features. The encoder refines Gaussian properties based on the updated features.

Our main contributions are summarized as follows:

\begin{itemize}[leftmargin=*]\setlength{\itemsep}{0.18em}
    \item We propose GaussianMap, a multi-sensor online HD map construction framework that represents the surrounding scene with adaptive Gaussian primitives on the BEV plane.
    \item We design a Gaussian encoder that generates Gaussian representations from sensor observations, with Gaussian self-attention modeling interactions among Gaussians.
    \item We formulate multi-sensor fusion in the Gaussian representation space and introduce Gaussian-LiDAR attention to incorporate geometric cues from LiDAR BEV features.
    \item Extensive experiments on nuScenes and Argoverse~2 show that GaussianMap achieves state-of-the-art performance in both camera-only and camera-LiDAR fusion settings.
\end{itemize}

%% file: figures/3_teaser.tex
\begin{figure}[t!] 
  \centering
  \includegraphics[width=\columnwidth]{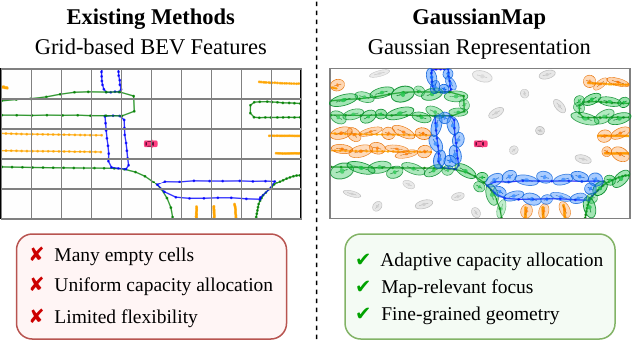}
  \caption{\small 
  \textbf{Motivation of GaussianMap.}
  Compared with grid-based BEV features that uniformly allocate representational capacity, GaussianMap represents the surrounding scene with adaptive Gaussian primitives to focus on map-relevant regions and perseve fine-grained geometry. Gaussian colors are used only for visualization.
  }
  \label{fig:teaser}
  \vspace{-4mm}
\end{figure}


%% file: 2_related_work.tex
\section{Related Work}  \label{sec:related_work}

\subsection{Online HD Map Construction}
Online HD map construction has attracted increasing attention for building local map representations from onboard sensor observations. Early methods~\cite{philion2020lift, li2024bevformer, liu2023bevfusion} mainly rely on BEV semantic segmentation to predict rasterized map layouts, while HDMapNet~\cite{li2022hdmapnet} further groups segmentation results into vectorized instances. Recent works have shifted toward end-to-end vectorized map construction. VectorMapNet~\cite{liu2023vectormapnet} introduces a two-stage auto-regressive framework for sequential point prediction, while the MapTR series~\cite{liao2022maptr, liao2024maptrv2} formulates this task as one-stage point-set prediction with permutation-equivalent modeling. BeMapNet~\cite{qiao2023end} and PivotNet~\cite{ding2023pivotnet} explore piecewise Bézier curves and pivot points to represent map elements. Other works improve vectorized map construction from complementary perspectives, including temporal modeling~\cite{yuan2024streammapnet}, mask-guided learning~\cite{liu2024mgmap}, and diffusion-based refinement~\cite{gao2026refdiffmap}. MapQR~\cite{liu2024leveraging} introduces scatter-and-gather queries to mitigate content conflicts among points of the same map element, while ADMap~\cite{hu2024admap} models point-sequence relationships within and across map elements using a cascading design. DAMap~\cite{dong2025damap} investigates classification-localization misalignment to improve prediction consistency. While existing methods commonly adopt grid-based BEV features, GaussianMap represents the surrounding scene with adaptive Gaussian primitives on the BEV plane to focus on map-relevant regions.


\subsection{Gaussian Representation}
Gaussian representation has attracted increasing attention as an efficient explicit approach for 3D scene modeling. 3D Gaussian Splatting~\cite{kerbl20233d} represents a scene as a set of 3D Gaussians and enables real-time radiance field rendering via differentiable splatting. Compared with voxel grids and neural radiance fields~\cite{mildenhall2021nerf}, Gaussian primitives provide an explicit and adaptive representation that concentrates modeling capacity around spatial structures. This representation has been extended to dynamic scene reconstruction~\cite{wu20244d}, 3D content generation~\cite{tang2024dreamgaussian}, and semantic scene understanding~\cite{qin2024langsplat}. Recent works have explored Gaussian representation for autonomous driving. The GaussianFormer series~\cite{huang2024gaussianformer, huang2025gaussianformer} adopts sparse 3D semantic Gaussians for 3D occupancy prediction, reducing redundancy in dense voxel representations. GaussianBeV~\cite{chabot2025gaussianbev} predicts 3D Gaussians from multi-view images for BEV semantic segmentation, while GaussianLSS~\cite{lu2025toward} models depth uncertainty with 3D Gaussians for BEV feature aggregation. GaussianAD~\cite{zheng2024gaussianad} models future scene evolution with 3D Gaussian flows for end-to-end autonomous driving. While most existing methods rely on 3D Gaussians, GaussianMap generates Gaussian primitives on the BEV plane in a feed-forward manner, making it well suited for online HD map construction.

\subsection{Multi-Sensor Fusion}
Multi-sensor fusion has attracted increasing attention in autonomous driving perception for integrating complementary information from different sensors. Existing methods can be broadly categorized into point-level, proposal-level, and BEV-level fusion paradigms. Point-level fusion methods~\cite{vora2020pointpainting, wang2021pointaugmenting, yin2021multimodal} typically project image features or semantic predictions onto LiDAR points to augment point properties or generate denser point clouds. Proposal-level fusion methods~\cite{chen2017multi, nabati2021centerfusion, bai2022transfusion} usually associate object proposals with features from different modalities to enhance instance-level representations. BEV-level fusion methods~\cite{liang2022bevfusion, li2024bevformer, liu2023bevfusion} integrate multi-modal features within a unified BEV space to obtain structured scene representations. Recent works have explored multi-sensor fusion for online HD map construction. MapFusion~\cite{hao2025mapfusion} proposes cross-modal interaction and dual dynamic fusion modules to improve BEV feature fusion. RoboMap~\cite{hao2025really} addresses robustness under multi-sensor corruptions through data augmentation and modality dropout training. SEF-Map~\cite{fu2026sef} decomposes multi-modal BEV features into specialized subspaces and integrates them with uncertainty-aware expert fusion. Unlike these methods that mainly fuse dense BEV features, GaussianMap employs adaptive Gaussian primitives as geometry-aware queries to aggregate multi-sensor features, allowing more focused fusion for map-relevant regions.


\input{figures/4_overall}

%% file: figures/4_overall.tex
\begin{figure*}[t!] 
  \centering
  \includegraphics[width=\textwidth]{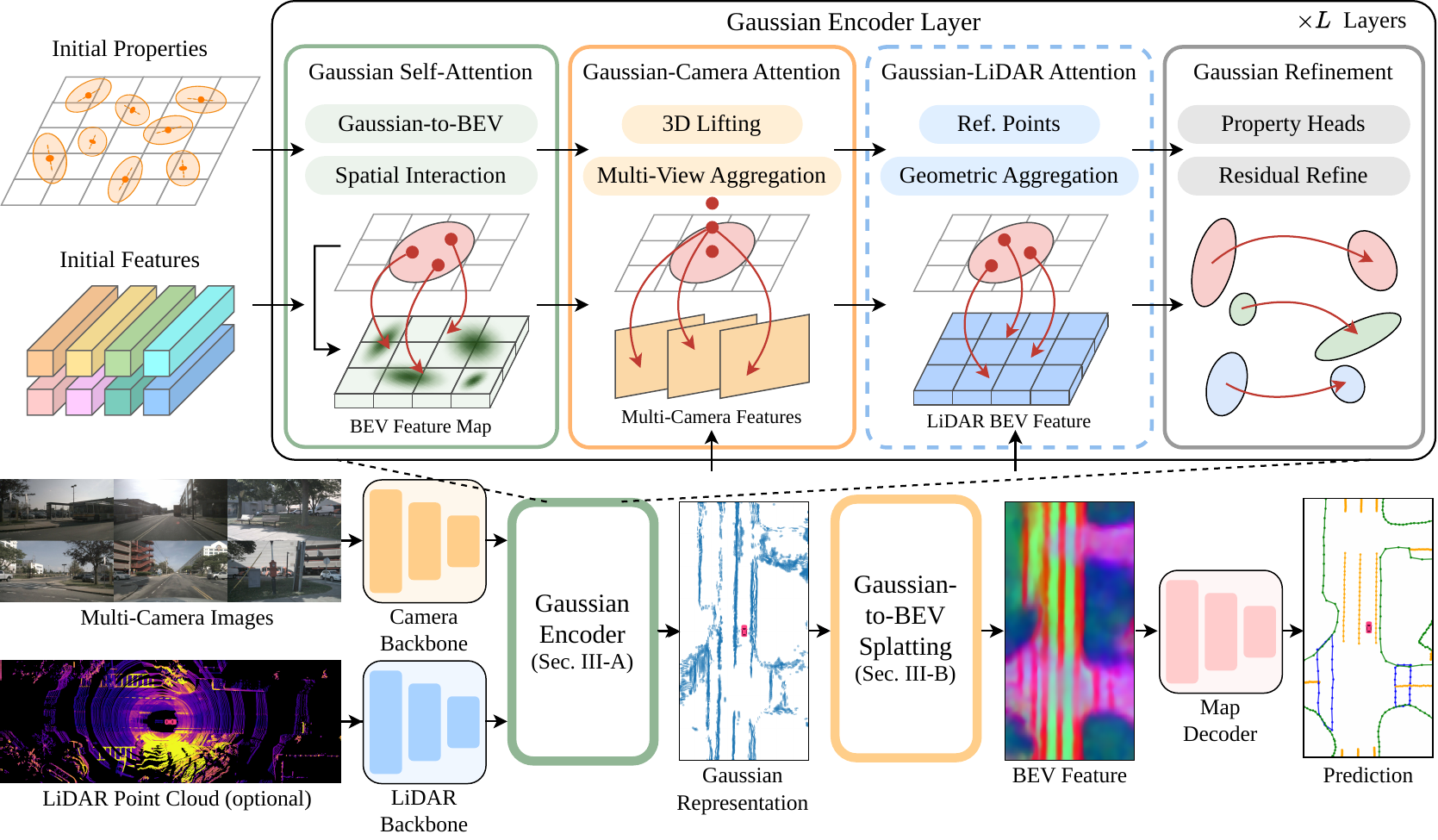}
  \caption{\small 
  \textbf{Overall framework of GaussianMap.} 
  The framework takes multi-camera images and an optional LiDAR point cloud as inputs. The Gaussian encoder generates an adaptive Gaussian representation on the BEV plane from the extracted sensor features. This representation is converted into a BEV feature map via Gaussian-to-BEV splatting, from which the map decoder produces the vectorized map prediction. For clarity, we visualize Gaussians with opacity higher than 0.4 and show the BEV feature map after principal component analysis.
  }
  \label{fig:overall_framework}
  \vspace{-4mm}
\end{figure*}

%% file: 3_method.tex
\section{Method} \label{sec:method}

GaussianMap constructs vectorized maps around the ego vehicle from multi-camera images and an optional LiDAR point cloud in an end-to-end manner. As illustrated in Fig.~\ref{fig:overall_framework}, the input images are first processed by a shared camera backbone to extract perspective-view features $\mathbf{F}_{\mathrm{cam}}^{\mathrm{PV}}$. When LiDAR is available, the point cloud is encoded by a LiDAR backbone into BEV features $\mathbf{F}_{\mathrm{lidar}}^{\mathrm{BEV}}$. These sensor features are fed into the Gaussian encoder (Sec.~\ref{sec:gaussian_encoder}) to generate a Gaussian representation \(\mathcal{G}=\{g_i\}_{i=1}^{M}\) on the BEV plane. Each primitive \(g_i\) is parameterized by a mean position \(\mathbf{m}_i\), scale \(\mathbf{s}_i\), rotation \(\mathbf{r}_i\), opacity \(\alpha_i\), and feature vector \(\mathbf{f}_i\). The Gaussian
representation is then converted into a BEV feature map $\mathbf{B}$ via Gaussian-to-BEV splatting (Sec.~\ref{sec:splatting}). Finally, a map decoder takes $\mathbf{B}$ as input and predicts the vectorized map $\hat{\mathcal{Y}}=\{\hat{y}_k\}_{k=1}^{K}$, where each map element $\hat{y}_k$ is represented by a class label and a point sequence.

\subsection{Gaussian Encoder} \label{sec:gaussian_encoder}
The Gaussian encoder generates a Gaussian representation from the extracted sensor features. It first initializes Gaussian primitives on the BEV plane and iteratively refines them through \(L\) stacked layers. At each layer \(\ell\), Gaussian features are updated by Gaussian self-attention (GSA), Gaussian-camera attention (GCA), optional Gaussian-LiDAR attention (GLA), and a feed-forward network. The updated features are then used to refine the Gaussian properties.

\vspace{0.15cm}
\noindent \textbf{Gaussian Initialization.}
We initialize a set of $M$ Gaussian primitives on the BEV plane, denoted by $\mathcal{G}^{0}=\{g_i^{0}\}_{i=1}^{M}$. The mean positions are initialized uniformly over the BEV perception range, with fixed scales, identity rotations, and opacities set to $0.5$. To initialize the Gaussian
features, we lift camera features $\mathbf{F}_{\mathrm{cam}}^{\mathrm{PV}}$ to BEV space using an LSS-style transformation~\cite{philion2020lift}, obtaining $\mathbf{F}_{\mathrm{cam}}^{\mathrm{BEV}}$. This feature is used directly in the camera-only setting, or fused with LiDAR BEV features when LiDAR is available, yielding $\mathbf{F}_{\mathrm{init}}^{\mathrm{BEV}}$. Each Gaussian feature $\mathbf{f}_i^0$ is then obtained by bilinearly sampling $\mathbf{F}_{\mathrm{init}}^{\mathrm{BEV}}$ at its mean position $\mathbf{m}_i^0$.

\vspace{0.15cm}
\noindent \textbf{Gaussian Self-Attention.}
GSA models interactions among Gaussian primitives based on their spatial distributions. For each Gaussian $g_i$, we generate a set of reference points $\mathcal{P}_i=\{\mathbf{p}_{i,q}\}_{q=1}^{Q}$ within its local region to guide feature aggregation. Specifically, local offsets $\{\mathbf{o}_{i,q}\}_{q=1}^{Q}$ are predicted from $\mathbf{f}_i$ and transformed according to the geometry of $g_i$:
\begin{equation}
\mathbf{p}_{i,q} = \mathbf{m}_i +
\mathbf{R}_i \left( \mathbf{o}_{i,q} \odot \mathbf{s}_i \right),
\label{eq:transform}
\end{equation}
where $\mathbf{m}_i$ and $\mathbf{s}_i$ are the mean position and scale of $g_i$, and $\mathbf{R}_i$ is the rotation matrix derived from $\mathbf{r}_i$. To obtain the feature source, the Gaussian representation is splatted into a BEV feature map $\mathbf{F}_{\mathrm{self}}^{\mathrm{BEV}}$ using Gaussian-to-BEV splatting (Sec.~\ref{sec:splatting}). The Gaussian feature is then updated by
\begin{equation}
\mathbf{f}^{\mathrm{gsa}}_i =
\mathrm{DA}
\left(
\mathbf{f}_i, \mathcal{P}_i, \mathbf{F}_{\mathrm{self}}^{\mathrm{BEV}}
\right),
\end{equation}
where \(\mathbf{f}^{\mathrm{gsa}}_i\) denotes the updated feature of \(g_i\) after GSA, and $\mathrm{DA}(\cdot)$ denotes the deformable attention operation~\cite{zhu2020deformable}.

\vspace{0.15cm}
\noindent \textbf{Gaussian-Camera Attention.} 
GCA incorporates visual information from multi-camera image features into the Gaussian representation. For each Gaussian $g_i$, we lift its mean position to a set of 3D reference points $\mathcal{P}_i^{\mathrm{3d}}=\{\mathbf{p}_{i,h}^{\mathrm{3d}}\}_{h=1}^{H}$. Specifically, height offsets $\{\Delta z_{i,h}\}_{h=1}^{H}$ are predicted from the Gaussian feature and added to predefined height anchors:
\begin{equation}
\mathbf{p}_{i,h}^{\mathrm{3d}} =
[\mathbf{m}_i, z_h + \Delta z_{i,h}],
\end{equation}
where $z_h$ denotes the $h$-th height anchor. These reference points are then projected onto the image feature planes using camera intrinsics and extrinsics, yielding 2D reference locations $\mathcal{P}_{i,c}^{\mathrm{2d}}$ for camera $c$. The Gaussian feature is updated by aggregating image features from all camera views:
\begin{equation}
\mathbf{f}_i^{\mathrm{gca}} =
\frac{1}{|\mathcal{C}|}\sum_{c\in\mathcal{C}}
\mathrm{KA}\left(
\mathbf{f}^{\mathrm{gsa}}_i,
\mathcal{P}_{i,c}^{\mathrm{2d}},
\mathbf{F}_{\mathrm{cam},c}^{\mathrm{PV}}
\right),
\end{equation}
where \(\mathbf{f}^{\mathrm{gca}}_i\) denotes the updated feature of \(g_i\) after GCA, and $\mathrm{KA}(\cdot)$ denotes the kernel-based attention operation~\cite{chen2022efficient}. 

\vspace{0.15cm}
\noindent \textbf{Gaussian-LiDAR Attention.} 
When LiDAR is available, we introduce GLA to incorporate complementary geometric cues into the Gaussian representation. GLA uses each Gaussian primitive as an adaptive query to aggregate LiDAR features from its spatially relevant region. For each Gaussian $g_i$, we generate another set of reference points 
$\mathcal{P}'_i$ using the same transformation as in Eq.~\ref{eq:transform}. 
These reference points guide feature aggregation on the LiDAR BEV feature map $\mathbf{F}_{\mathrm{lidar}}^{\mathrm{BEV}}$. The Gaussian feature is then updated by
\begin{equation}
\mathbf{f}^{\mathrm{gla}}_i =
\mathrm{DA}
\left(
\mathbf{f}^{\mathrm{gca}}_i,
\mathcal{P}'_i,
\mathbf{F}_{\mathrm{lidar}}^{\mathrm{BEV}}
\right),
\end{equation}
where \(\mathbf{f}^{\mathrm{gla}}_i\) denotes the updated feature of \(g_i\) after GLA.

\vspace{0.15cm}
\noindent \textbf{Gaussian Refinement.}
After feature interaction in each encoder layer, we refine the Gaussian properties based on the updated features. For the $i$-th Gaussian primitive 
$g_i=(\mathbf{m}_i,\mathbf{s}_i,\mathbf{r}_i,\alpha_i,\mathbf{f}_i)$,
its updated feature $\mathbf{f}'_i$ encodes contextual information from Gaussian interactions and sensor features. We use separate multi-layer perceptron (MLP) heads to predict intermediate Gaussian properties:
\begin{equation}
\begin{alignedat}{2}
\mathbf{m}'_i     & = \mathrm{MLP}_m(\mathbf{f}'_i), \quad&
\mathbf{s}'_i     & = \mathrm{MLP}_s(\mathbf{f}'_i), \\
\mathbf{r}'_i     & = \mathrm{MLP}_r(\mathbf{f}'_i), \quad&
\alpha'_i         & = \mathrm{MLP}_\alpha(\mathbf{f}'_i).
\end{alignedat}
\end{equation}
We then update the mean with the predicted residual $\mathbf{m}'_i$, while directly replacing the remaining properties:
\begin{equation}
g'_i
=
\left(
\mathbf{m}_i + \mathbf{m}'_i,
\mathbf{s}'_i,
\mathbf{r}'_i,
\alpha'_i,
\mathbf{f}'_i
\right).
\end{equation}
The residual mean update allows Gaussians to progressively adjust their positions and stabilizes the refinement process.

\subsection{Gaussian-to-BEV Splatting} \label{sec:splatting}

Gaussian primitives provide an adaptive scene representation on the BEV plane, while many feature aggregation operators, such as deformable attention~\cite{zhu2020deformable} and convolution, require structured feature maps. We therefore introduce Gaussian-to-BEV splatting to transform the Gaussian representation into a structured BEV feature map. Specifically, we model Gaussian primitives as mixture components and aggregate their features using posterior responsibilities.

For a BEV cell centered at location $\mathbf{x}$, we evaluate the conditional probability density of $\mathbf{x}$ under the $i$-th Gaussian primitive $g_i=(\mathbf{m}_i,\mathbf{s}_i,\mathbf{r}_i,\alpha_i,\mathbf{f}_i)$ as
\begin{equation}
p(\mathbf{x}\mid g_i)
=
\frac{1}{2\pi |\boldsymbol{\Sigma}_i|^{\frac{1}{2}}}
\exp\bigl(
-\frac{1}{2}
(\mathbf{x}-\mathbf{m}_i)^{\top}
\boldsymbol{\Sigma}_i^{-1}
(\mathbf{x}-\mathbf{m}_i)
\bigr)
\end{equation}
where $\boldsymbol{\Sigma}_i$ is the covariance matrix derived from $\mathbf{s}_i$ and $\mathbf{r}_i$. We use the opacity $\alpha_i$ as the mixture prior and compute the posterior responsibility of $g_i$ for location $\mathbf{x}$ as
\begin{equation}
p(g_i\mid \mathbf{x})
=
\frac{
p(\mathbf{x}\mid g_i)\alpha_i
}{
\sum_{j=1}^{M} p(\mathbf{x}\mid g_j)\alpha_j
}.
\end{equation}
The BEV feature $\mathbf{B}$ at $\mathbf{x}$ is then obtained as the expectation of Gaussian features under this posterior distribution:
\begin{equation}
\mathbf{B}(\mathbf{x})
=
\sum_{i=1}^{M}
p(g_i\mid \mathbf{x}) \mathbf{f}_i .
\end{equation}
This operation is implemented with a CUDA-based splatter that evaluates Gaussian
contributions at BEV cell centers.

\subsection{Vectorized Map Prediction and Supervision}

After Gaussian-to-BEV splatting, we feed the resulting BEV feature \(\mathbf{B}\) into a vectorized map decoder to produce structured map elements. Since \(\mathbf{B}\) follows the standard BEV feature format, it is compatible with existing vectorized map decoders, such as MapTR/MapTRv2 \cite{liao2022maptr,liao2024maptrv2}, MapQR~\cite{liu2024leveraging}, and DAMap~\cite{dong2025damap}. Consequently, the proposed Gaussian representation does not require a specialized map decoder. Instead, it provides an enhanced scene representation for vectorized map prediction, as evaluated in Sec.~\ref{sec:ablation}.

The training objective follows common practice in map construction \cite{liao2024maptrv2,liu2024leveraging,dong2025damap} and consists of a vectorized map prediction loss and a dense auxiliary loss:
\begin{equation}
\mathcal{L}
=
\lambda_{\mathrm{map}}\mathcal{L}_{\mathrm{map}}
+
\lambda_{\mathrm{dense}}\mathcal{L}_{\mathrm{dense}},
\end{equation}
where \(\lambda_{\mathrm{map}}\) and \(\lambda_{\mathrm{dense}}\) balance the two terms. The map loss \(\mathcal{L}_{\mathrm{map}}\) includes classification, point-to-point, and edge direction losses, while the dense loss \(\mathcal{L}_{\mathrm{dense}}\) includes depth, BEV segmentation, and perspective-view segmentation losses.

%% file: 4_experiments.tex
\section{Experiments} \label{sec:experiments}

\subsection{Experimental Setup}

\noindent \textbf{Datasets.}
We evaluate our method on two public real-world benchmarks, nuScenes~\cite{caesar2020nuscenes} and Argoverse~2~\cite{wilson2023argoverse}. nuScenes contains 1,000 driving scenes recorded in Boston and Singapore, each lasting about \SI{20}{\second}. Each sample provides images from six surrounding cameras, point clouds from a 32-beam LiDAR, and vector maps annotated at \SI{2}{\hertz}. Argoverse~2 consists of 1,000 driving logs collected in six U.S. cities, with each log lasting about \SI{15}{\second}. Each sample includes images from seven surrounding cameras, point clouds from two 32-beam LiDARs, and vector maps annotated at \SI{10}{\hertz}. Following previous works~\cite{liao2024maptrv2, liu2024leveraging}, we use 700 scenes for training and 150 scenes for validation on both datasets. We report performance on three categories, including lane dividers, pedestrian crossings, and road boundaries.

\vspace{0.15cm}
\noindent \textbf{Evaluation Metrics.}
For a fair comparison, we follow the standard evaluation protocol used in previous works~\cite{liao2024maptrv2, liu2024leveraging}. The perception range is centered on the ego vehicle and is set to $[\SI{-15}{\meter}, \SI{15}{\meter}]$ along the X-axis and $[\SI{-30}{\meter}, \SI{30}{\meter}]$ along the Y-axis. We use mean Average Precision (mAP) to evaluate the quality of vectorized map predictions. Specifically, a predicted map element is considered a true positive if its Chamfer distance to a ground-truth instance is smaller than a given threshold. For each category, AP is computed under three thresholds $\{\SI{0.5}{\meter}, \SI{1.0}{\meter}, \SI{1.5}{\meter}\}$. The final mAP is reported by averaging the AP scores across all categories. 

\vspace{0.15cm}
\noindent \textbf{Implementation Details.}
Unless otherwise specified, we follow the implementation settings used in previous works~\cite{liao2024maptrv2, liu2024leveraging}. We use ResNet-50~\cite{he2016deep} and SECOND~\cite{yan2018second} as the image and LiDAR backbones, respectively. For a fair comparison, we set the number of Gaussians to 20,000, which is the same as the number of cells in the BEV feature of size $200 \times 100$. The Gaussian encoder refines the features and properties of these Gaussians using 3 encoder layers. We adopt the MapQR$+$DAMap~\cite{dong2025damap} decoder with 6 decoder layers and 100 instance queries. Each map element is represented by a class label and a sequence of 20 points. Our model is trained on 8 NVIDIA GPUs with a batch size of 32. The training schedule is 24 epochs on nuScenes and 6 epochs on Argoverse~2. We use the AdamW optimizer with a learning rate of $6 \times 10^{-4}$ and a cosine annealing schedule.

\subsection{Comparisons with State-of-the-art Methods}

\input{tables/1_nus_results}

\noindent \textbf{Results on nuScenes.}
Table~\ref{tab:nus_results} compares GaussianMap with state-of-the-art methods on the nuScenes validation set. In the camera-only setting, GaussianMap achieves 70.5~mAP with a 24-epoch training schedule, surpassing the previous best camera-based method MapQR$+$DAMap~\cite{dong2025damap} by 1.7~mAP. In the camera-LiDAR fusion setting, GaussianMap achieves 78.3~mAP under the same training schedule, outperforming RoboMap~\cite{hao2025really} and the strong MapQR$+$DAMap~\cite{dong2025damap} baseline by 1.3 and 3.1~mAP, respectively. These consistent improvements under both input settings suggest that the proposed Gaussian representation serves as an effective intermediate scene representation for vectorized map prediction.

\input{tables/2_av2_results}

\input{tables/3_ablate_encoder}

\vspace{0.15cm}
\noindent \textbf{Results on Argoverse~2.}
Table~\ref{tab:av2_results} compares GaussianMap with state-of-the-art methods on the Argoverse~2 validation set. In the camera-only setting, GaussianMap achieves 70.1~mAP with a 6-epoch training schedule, outperforming RelMap~\cite{cai2025relmap} and the strong MapQR$+$DAMap~\cite{dong2025damap} baseline by 0.2 and 2.0~mAP, respectively. In the camera-LiDAR fusion setting, GaussianMap achieves 79.0~mAP under the same training schedule, surpassing ADMap~\cite{hu2024admap} and MapQR$+$DAMap~\cite{dong2025damap} by 2.1 and 2.9~mAP, respectively. These consistent improvements on Argoverse~2 further demonstrate the effectiveness of the proposed Gaussian representation across different benchmarks and input settings.

\subsection{Ablation Study} \label{sec:ablation} 
We conduct ablation studies on the nuScenes dataset to evaluate the effectiveness of the proposed GaussianMap. All experiments are conducted for 24 epochs. Unless otherwise specified, we adopt MapQR~\cite{liu2024leveraging} as the baseline method.

\input{tables/4_ablate_main}

\vspace{0.15cm}
\noindent \textbf{Effect of Gaussian Encoder.}
To validate the effectiveness of the proposed Gaussian encoder, we integrate it into two baselines, MapQR~\cite{liu2024leveraging} and MapQR$+$DAMap~\cite{dong2025damap}. As shown in Table~\ref{tab:ablate_encoder}, the Gaussian encoder consistently improves both baselines across different input settings, indicating that the Gaussian representation generated by the encoder benefits vectorized map prediction. In the camera-only setting, it improves MapQR and MapQR$+$DAMap by 2.1 and 1.7~mAP, respectively, suggesting that refining Gaussian primitives from visual features allows more effective capacity allocation to map-relevant regions. In the camera-LiDAR fusion setting, the gains increase to 4.9 and 3.1~mAP, respectively, suggesting that the encoder further enhances the Gaussian representation with complementary geometric cues.

\vspace{0.15cm}
\noindent \textbf{Contributions of Main Components.} 
To analyze the contribution of each component in the Gaussian encoder, we progressively add Gaussian representation, Gaussian self-attention (GSA), and Gaussian-LiDAR attention (GLA) to the baseline, as reported in Table~\ref{tab:ablate_main}. In the camera-only setting, introducing the Gaussian representation brings a 1.5~mAP gain, suggesting that adaptive Gaussian primitives provide a more effective representation for map-relevant regions. Building on this representation, GSA further increases performance to 68.5~mAP by modeling interactions among Gaussians. In the camera-LiDAR fusion setting, Gaussian representation and GSA together bring a 2.5~mAP gain over the baseline. Adding GLA brings an additional gain of 2.4~mAP, suggesting that GLA effectively incorporates LiDAR geometric cues into the Gaussian representation.

\vspace{0.15cm}
\noindent \textbf{Ablation on Gaussian-Guided Feature Aggregation.} 
To evaluate the effect of Gaussian-guided feature aggregation in GSA and GLA, we vary the number of reference points used by each Gaussian primitive. As shown in Table~\ref{tab:ablate_sampling}, increasing the number of reference points from 4 to 16 improves the mAP from 76.9 to 77.8, suggesting that denser reference points help each Gaussian capture richer spatial context. Further increasing the number to 32 yields a comparable mAP of 77.7, indicating that the benefit tends to saturate once the reference points become sufficiently dense.

\input{tables/5_ablate_sample}

\input{tables/6_ablate_fusion}

\vspace{0.15cm}
\noindent \textbf{Ablation on Multi-Sensor Fusion Mechanisms.} 
To evaluate the effectiveness of Gaussian-LiDAR attention, we compare it with two alternative multi-sensor fusion mechanisms: convolutional fusion following~\cite{liao2024maptrv2} and deformable attention~\cite{zhu2020deformable}. As shown in Table~\ref{tab:ablate_fusion}, convolutional fusion and deformable attention achieve 75.4 and 77.3~mAP, respectively. Gaussian-LiDAR attention achieves the best performance, outperforming the two alternatives by 2.4 and 0.5~mAP, respectively. This suggests that using Gaussian primitives as spatially adaptive queries allows more effective aggregation of LiDAR geometric cues into the Gaussian representation.

\input{figures/1_qualitative_results}

\subsection{Qualitative Results}
We present qualitative comparisons on the nuScenes~\cite{caesar2020nuscenes} dataset in both the camera-only and camera-LiDAR fusion settings, as shown in Fig.~\ref{fig:qualitative_results}(a) and (b), respectively. We also visualize Gaussians with opacity higher than 0.4 to highlight regions with stronger responses. Compared with MapQR~\cite{liu2024leveraging} and MapQR$+$DAMap~\cite{dong2025damap}, GaussianMap produces vectorized maps that are better aligned with the ground truth across diverse driving scenes. The Gaussians tend to concentrate in map-relevant regions, suggesting that the proposed Gaussian representation can adaptively allocate modeling capacity to geometrically informative regions. We further observe that incorporating LiDAR geometric cues yields more compact and better localized Gaussians around map-relevant regions.

%% file: tables/1_nus_results.tex
\begin{table}[t!]
\centering
\caption{
\small 
\textbf{Comparison with state-of-the-art methods on the nuScenes validation set.} 
“C” and “L” denote camera and LiDAR inputs, respectively. $\dagger$ indicates results reproduced with the official code; for camera-LiDAR fusion, we follow the mechanism in \cite{liao2024maptrv2}. Results of other methods are taken from the corresponding papers.
}

{
\setlength{\tabcolsep}{5pt}
\begin{tabularx}{\columnwidth}{@{}>{\raggedright\arraybackslash}X@{\hspace{3pt}}|c@{\hspace{4pt}}c|cccc@{}}
\toprule
\multicolumn{1}{c|}{\multirow{2}{*}{Method}} 
& \multirow{2}{*}{Modality} 
& \multirow{2}{*}{Epoch} 
& \multicolumn{4}{c}{ AP } \\
& & 
& div. & ped. & bou. & mean \\
\midrule
HDMapNet \cite{li2022hdmapnet}                  & C & 30 & 21.7 & 14.4 & 33.0 & 23.0 \\
VectorMapNet \cite{liu2023vectormapnet}         & C & 110 & 51.4 & 42.5 & 44.1 & 46.0 \\
MapTR \cite{liao2022maptr}                      & C & 24 & 51.5 & 46.3 & 53.1 & 50.3 \\
MapVR \cite{zhang2024online}                    & C & 24 & 54.4 & 47.7 & 51.4 & 51.2 \\
PivotNet \cite{ding2023pivotnet}                & C & 24 & 56.5 & 56.2 & 60.1 & 57.6 \\
BeMapNet \cite{qiao2023end}                     & C & 30 & 62.3 & 57.7 & 59.4 & 59.8 \\
MapTRv2 \cite{liao2024maptrv2}                  & C & 24 & 62.4 & 59.8 & 62.4 & 61.5 \\
HeightMapNet \cite{qiu2025heightmapnet}         & C & 24 & 62.6 & 60.4 & 62.2 & 61.7 \\
ADMap \cite{hu2024admap}                        & C & 24 & 63.5 & 61.9 & 63.3 & 62.9 \\
RefDiffMap \cite{gao2026refdiffmap}             & C & 24 & 64.0 & 64.6 & 64.3 & 64.3 \\
MGMap \cite{liu2024mgmap}                       & C & 24 & 65.0 & 61.8 & 67.5 & 64.8 \\
MapQR \cite{liu2024leveraging}                  & C & 24 & 68.0 & 63.4 & 67.7 & 66.4 \\
HIMap \cite{zhou2024himap}                      & C & 30 & 68.4 & 62.6 & 69.1 & 66.7 \\
RelMap \cite{cai2025relmap}                     & C & 24 & 70.5 & 66.3 & 68.4 & 68.4 \\
MapQR$+$DAMap \cite{dong2025damap}              & C & 24 & 70.8 & 65.2 & 70.3 & 68.8 \\
\textbf{GaussianMap (Ours)}                     & C & 24 & 72.2 & 68.1 & 71.3 & \textbf{70.5} \\

\midrule
HDMapNet \cite{li2022hdmapnet}                  & C \& L & 30 & 29.6 & 16.3 & 46.7 & 31.0 \\
VectorMapNet \cite{liu2023vectormapnet}         & C \& L & 110 & 60.1 & 48.2 & 53.0 & 53.7 \\
MapTR \cite{liao2022maptr}                      & C \& L & 24 & 62.3 & 55.9 & 69.3 & 62.5 \\
MapVR \cite{zhang2024online}                    & C \& L & 24 & 62.7 & 60.4 & 67.2 & 63.5 \\
MapFusion \cite{hao2025mapfusion}               & C \& L & 24 & 64.4 & 61.6 & 72.5 & 66.1 \\
SEF-Map \cite{fu2026sef}                        & C \& L & 24 & 66.7 & 61.6 & 71.8 & 66.7 \\
MapTRv2 \cite{liao2024maptrv2}                  & C \& L & 24 & 66.5 & 65.6 & 74.8 & 69.0 \\
GeMap \cite{gemap}                              & C \& L & 110 & 69.8 & 68.0 & 73.4 & 70.4 \\
ADMap \cite{hu2024admap}                        & C \& L & 24 & 69.0 & 68.0 & 75.2 & 70.8 \\
MGMap \cite{liu2024mgmap}                       & C \& L & 24 & 71.1 & 67.7 & 76.2 & 71.7 \\
MapQR$^{\dagger} $\cite{liu2024leveraging}      & C \& L & 24 & 71.0 & 70.6 & 77.1 & 72.9 \\
HIMap \cite{zhou2024himap}                      & C \& L & 24 & 72.4 & 71.0 & 79.4 & 74.3 \\
MapQR$+$DAMap$^{\dagger} $\cite{dong2025damap}  & C \& L & 24 & 75.6 & 71.6 & 78.4 & 75.2 \\
RoboMap \cite{hao2025really}                    & C \& L & 24 & 74.5 & 74.6 & 82.0 & 77.0 \\
\textbf{GaussianMap (Ours)}                     & C \& L & 24 & 78.1 & 74.4 & 82.3 & \textbf{78.3} \\
\bottomrule
\end{tabularx}
}

\label{tab:nus_results} 
\vspace{-4mm}
\end{table}




%% file: tables/2_av2_results.tex
\begin{table}[t!]
\centering
\caption{
\small 
\textbf{Comparison with state-of-the-art methods on the Argoverse~2 validation set.} 
“C” and “L” denote camera and LiDAR inputs, respectively. $\dagger$ indicates results reproduced with the official code; for camera-LiDAR fusion, we follow the mechanism in \cite{liao2024maptrv2}. Results of other methods are taken from the corresponding papers.
}

{
\setlength{\tabcolsep}{5pt}
\begin{tabularx}{\columnwidth}{@{}>{\raggedright\arraybackslash}X@{\hspace{3pt}}|c@{\hspace{4pt}}c|cccc@{}}
\toprule
\multicolumn{1}{c|}{\multirow{2}{*}{Method}} 
& \multirow{2}{*}{Modality} 
& \multirow{2}{*}{Epoch} 
& \multicolumn{4}{c}{ AP } \\
& & 
& div. & ped. & bou. & mean \\
\midrule
HDMapNet \cite{li2022hdmapnet}                    & C & - & 5.7  & 13.1 & 37.6 & 18.8 \\
VectorMapNet \cite{liu2023vectormapnet}           & C & - & 36.1 & 38.3 & 39.2 & 37.9 \\
MapVR \cite{zhang2024online}                      & C & 6 & 60.0 & 54.6 & 58.0 & 57.5 \\
MapTR \cite{liao2022maptr}                        & C & 6 & 65.5 & 56.6 & 61.8 & 61.3 \\
HeightMapNet \cite{qiu2025heightmapnet}           & C & 6 & 70.6 & 67.9 & 62.7 & 67.1 \\
MapTRv2 \cite{liao2024maptrv2}                    & C & 6 & 72.1 & 62.9 & 67.1 & 67.4 \\
GeMap \cite{gemap}                                & C & 6 & 67.6 & 59.3 & 64.7 & 63.9 \\
MapQR$+$DAMap$^{\dagger} $\cite{dong2025damap}    & C & 6 & 72.2 & 63.9 & 68.3 & 68.1 \\
MapQR \cite{liu2024leveraging}                    & C & 6 & 72.3 & 64.3 & 68.1 & 68.2 \\
ADMap \cite{hu2024admap}                          & C & 6 & 72.4 & 64.5 & 68.9 & 68.7 \\
HIMap \cite{zhou2024himap}                        & C & 6 & 69.5 & 69.0 & 70.3 & 69.6 \\
RelMap \cite{cai2025relmap}                       & C & 6 & 74.2 & 65.3 & 70.1 & 69.9 \\
\textbf{GaussianMap (Ours)}                       & C & 6 & 73.1 & 65.2 & 71.9 & \textbf{70.1} \\

\midrule
MapTR \cite{liao2022maptr}                        & C \& L & 6 & 65.1 & 61.6 & 75.1 & 67.3 \\
MapFusion \cite{hao2025mapfusion}                 & C \& L & 6 & 69.4 & 65.8 & 78.9 & 71.4 \\
SEF-Map \cite{fu2026sef}                          & C \& L & 6 & 66.7 & 70.7 & 78.8 & 72.1 \\
MapQR$+$DAMap$^{\dagger} $\cite{dong2025damap}    & C \& L & 6 & 77.0 & 70.5 & 80.9	& 76.1 \\
ADMap \cite{hu2024admap}                          & C \& L & 6 & 76.2 & 72.8 & 81.5 & 76.9 \\
\textbf{GaussianMap (Ours)}                       & C \& L & 6 & 79.4 &	73.7 & 83.8 & \textbf{79.0} \\

\bottomrule
\end{tabularx}
}

\label{tab:av2_results} 
\vspace{-1mm}
\end{table}

%% file: tables/3_ablate_encoder.tex
\begin{table}[t!]
\centering
\caption{
\small 
\textbf{Integrating the Gaussian encoder into other methods.}
“C” and “L” denote camera and LiDAR input modalities, respectively. $\dagger$ indicates results reproduced with the official code; for camera-LiDAR fusion,  we follow the mechanism used in \cite{liao2024maptrv2}. The best results under the same settings are highlighted in bold.
}

\begin{tabularx}{0.96\columnwidth}{@{}>{\raggedright\arraybackslash}X@{\hspace{3pt}}|c|ccc@{\hspace{9pt}}c@{}}
\toprule
\multicolumn{1}{c|}{\multirow{2}{*}{Method}} & \multirow{2}{*}{Modality} & \multicolumn{4}{c}{AP} \\
 &  & div. & ped. & bou. & mean \\
\midrule
\multirow{2}{*}{MapQR$^{\dagger}$} 
    & C      & 68.0 & 63.4 & 67.7 & 66.4 \\
    & C \& L & 71.0 & 70.6 & 77.1 & 72.9 \\
\multirow{2}{*}{MapQR$+$Ours}      
    & C      & 70.9 & 65.0 & 69.5 & \textbf{68.5}$^{\textcolor{red}{\uparrow2.1}}$ \\
    & C \& L & 77.9 & 75.5 & 79.9 & \textbf{77.8}$^{\textcolor{red}{\uparrow4.9}}$ \\
    
\midrule
\multirow{2}{*}{MapQR$+$DAMap$^{\dagger}$} 
    & C      & 70.8 & 65.2 & 70.3 & 68.8 \\
    & C \& L & 75.6 & 71.6 & 78.4 & 75.2 \\
\multirow{2}{*}{MapQR$+$DAMap$+$Ours}      
    & C      & 72.2 & 68.1 & 71.3 & \textbf{70.5}$^{\textcolor{red}{\uparrow1.7}}$ \\
    & C \& L & 78.1 & 74.4 & 82.3 & \textbf{78.3}$^{\textcolor{red}{\uparrow3.1}}$\\

\bottomrule
\end{tabularx}

\label{tab:ablate_encoder} 
\vspace{-4mm}
\end{table}

%% file: tables/4_ablate_main.tex
\begin{table}[t!]
\centering
\caption{
\small 
\textbf{Ablation study of main components in the Gaussian encoder.} 
“C” and “L” denote camera and LiDAR inputs, respectively. “Gauss. Rep.”, “GSA”, and “GLA” denote Gaussian representation, Gaussian self-attention, and Gaussian-LiDAR attention. Best results under the same setting are highlighted in bold.
}

\begin{tabularx}{0.96\columnwidth}{>{\centering\arraybackslash}X|ccc|cccc@{}}
\toprule
\multirow{2}{*}{Modality} 
& \multirow{2}{*}{\makecell{Gauss.\\Rep.}} 
& \multirow{2}{*}{GSA} 
& \multirow{2}{*}{GLA} 
& \multicolumn{4}{c}{AP} \\
 &  &  &  & div. & ped. & bou. & mean \\
\midrule

\multirow{3}{*}{C} 
&           &           &           & 68.0 & 63.4 & 67.7 & 66.4 \\ 
& \ding{51} &           &           & 69.8 & 64.7 & 69.1 & 67.9 \\ 
& \ding{51} & \ding{51} &           & 70.9 & 65.0 & 69.5 & \textbf{68.5} \\ 

\midrule
\multirow{3}{*}{C \& L} 
&           &           &           & 71.0 & 70.6 & 77.1 & 72.9 \\ 
& \ding{51} & \ding{51} &           & 74.9 & 73.3 & 78.0 & 75.4 \\ 
& \ding{51} & \ding{51} & \ding{51} & 77.9 & 75.5 & 79.9 & \textbf{77.8} \\ 

\bottomrule
\end{tabularx}

\label{tab:ablate_main} 
\vspace{-4mm}
\end{table}

%% file: tables/5_ablate_sample.tex
\begin{table}[t!]
\centering
\caption{
\small 
\textbf{Ablation on Gaussian-guided feature aggregation.}
“Ref. Points per Gauss.” denotes the number of reference points used by each Gaussian for feature aggregation. “GSA” and “GLA” denote Gaussian self-attention and Gaussian-LiDAR attention.
}

\begin{tabularx}{0.83\columnwidth}{>{\centering\arraybackslash}X >{\centering\arraybackslash}X|cccc@{}}
\toprule
\multicolumn{2}{c|}{Ref. Points per Gauss.}
& \multicolumn{4}{c}{AP} \\
GSA & GLA & div. & ped. & bou. & mean \\

\midrule
4  & 4  & 76.2 & 75.2 & 79.3 & 76.9 \\
8  & 8  & 77.6 & 75.4 & 79.7 & 77.6 \\
16 & 16 & 77.9 & 75.5 & 79.9 & \textbf{77.8} \\
32 & 32 & 77.5 & 75.2 & 80.5 & 77.7 \\
\bottomrule
\end{tabularx}
\label{tab:ablate_sampling}
\vspace{-1mm}
\end{table}

%% file: tables/6_ablate_fusion.tex
\begin{table}[t!]
\centering
\caption{
\small 
\textbf{Ablation of multi-sensor fusion mechanisms.} 
“C” and “L” denote camera and LiDAR input modalities, respectively. Camera-only results are included for reference.
}

\begin{tabularx}{0.96\columnwidth}{@{}>{\raggedright\arraybackslash}X|@{\hspace{5pt}}c@{\hspace{5pt}}|cccc@{}}
\toprule
\multicolumn{1}{c@{\hspace{3pt}}|@{\hspace{5pt}}}{\multirow{2}{*}{Method}}
& \multirow{2}{*}{Modality}
& \multicolumn{4}{c}{AP} \\
& & div. & ped. & bou. & mean \\

\midrule
Camera Only                                     & C      & 70.9 & 65.0 & 69.5 & 68.5  \\
Convolutional Fusion \cite{liao2024maptrv2}     & C \& L & 74.9 & 73.3 & 78.0 & 75.4  \\
Deformable Attention \cite{zhu2020deformable}   & C \& L & 77.1 & 75.3 & 79.4 & 77.3  \\ 
Gaussian-LiDAR Attention                        & C \& L & 77.9 & 75.5 & 79.9 & \textbf{77.8}  \\

\bottomrule
\end{tabularx}

\label{tab:ablate_fusion} 
\vspace{-4mm}
\end{table}

%% file: figures/1_qualitative_results.tex
\begin{figure*}[t!] 
  \centering
  \includegraphics[width=\textwidth]{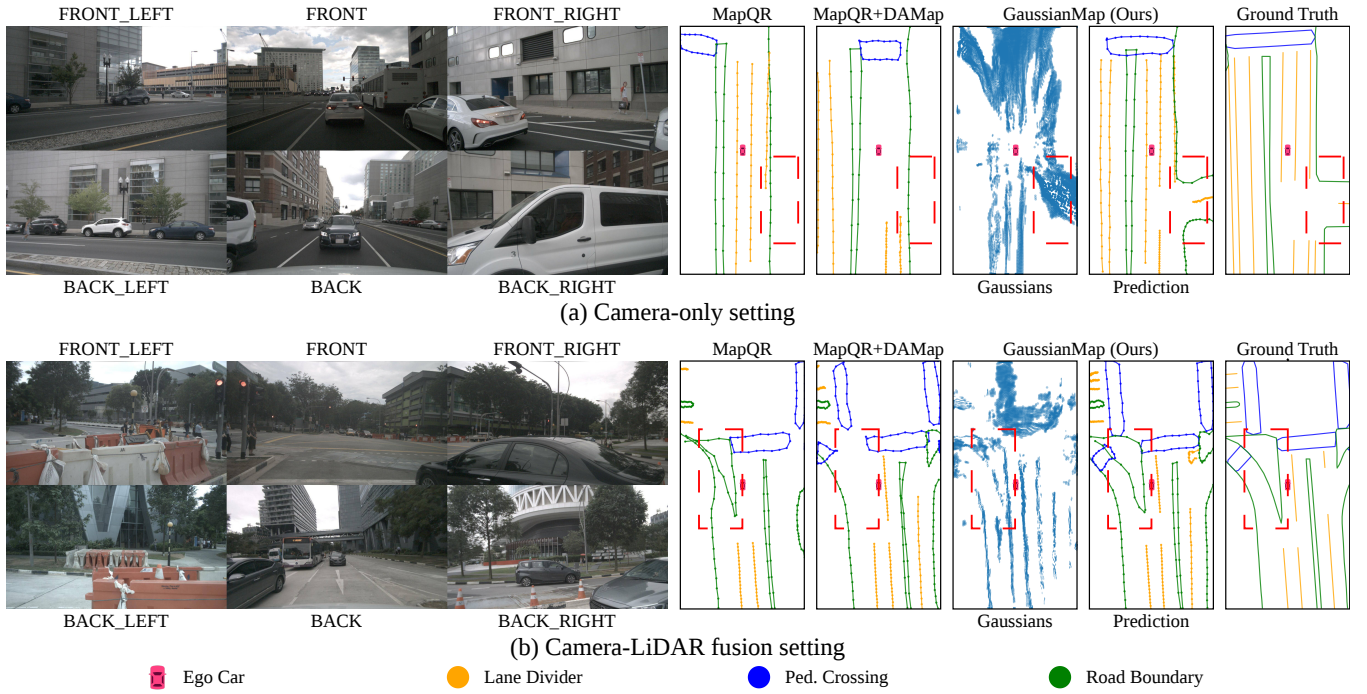}
  \caption{\small 
  \textbf{Qualitative results on the nuScenes validation set.} 
  (a) Camera-only setting. (b) Camera-LiDAR fusion setting.
  We compare the vectorized map predictions by GaussianMap with those from MapQR and MapQR$+$DAMap, using the ground truth as the reference. For GaussianMap, we additionally visualize the Gaussians alongside the predictions, retaining those with opacity higher than 0.4 for clarity.
  }
  \label{fig:qualitative_results}
  \vspace{-4mm}
\end{figure*}

%% file: 5_conclusion.tex
\section{Conclusion} \label{sec:conclusion}

In this paper, we present GaussianMap, a multi-sensor online HD map construction framework that learns an adaptive Gaussian representation of the surrounding scene. Built upon Gaussian primitives on the BEV plane, GaussianMap focuses modeling capacity on map-relevant regions and alleviates the redundancy of grid-based BEV features. A feed-forward Gaussian encoder refines these primitives through Gaussian interaction modeling and multi-sensor feature aggregation. Experiments on nuScenes and Argoverse~2 demonstrate that GaussianMap achieves state-of-the-art performance in both camera-only and camera-LiDAR settings. These results show that adaptive Gaussian representations provide an effective and flexible alternative for online HD map construction.